\documentclass[10pt]{article}
\usepackage[preprint]{tmlr}
\usepackage{microtype}
\usepackage{graphicx}
\usepackage{booktabs}
\usepackage{array}
\usepackage{multirow}
\usepackage{amsmath,amssymb}
\usepackage{algorithm}
\usepackage{algpseudocode}
\usepackage{nicefrac}
\usepackage{doi}
\usepackage{xcolor}
\usepackage{hyperref}
\hypersetup{
  colorlinks=true,
  allcolors=blue
}
\usepackage{url}
\usepackage{placeins}
\usepackage{adjustbox}
\usepackage{etoolbox}

\BeforeBeginEnvironment{tabular}{\begin{adjustbox}{max width=\linewidth}}
\AfterEndEnvironment{tabular}{\end{adjustbox}}
\AtBeginEnvironment{itemize}{\raggedright}

\title{\centering Fooling Reward Models with Gibberish: Adversarial Examples Beyond the Semantic Regime}
\renewcommand{\headrulewidth}{0pt}

\author{\centering\normalfont\fontsize{11}{14}\selectfont
Yuheng Zhang$^{1}$\thanks{Equal contribution. Email: \texttt{\{yuhengz2, mhuo5\}@illinois.edu}.},
Mingyue Huo$^{1}$\footnotemark[1], Minghao Zhu$^{2}$,
Mengxue Zhang$^{3}$, Nan Jiang$^{1}$ \\[0.6em]
{\normalsize $^{1}$UIUC \quad $^{2}$Independent Researcher} \\[0.3em]
{\normalsize $^{3}$University of Massachusetts Amherst}
\endgraf}

\newcommand{\TOMPAArxiv}{}

\begin{document}
\raggedbottom
\maketitle

\begin{abstract}
Reward models (RMs) are widely used as optimization targets in reinforcement learning from human feedback (RLHF). We show that leading RMs can assign higher scores to gibberish than to meaningful reference answers. To systematically discover such inputs, we introduce Token Mapping Perturbation Attack (TOMPA), a framework that uses a perturbed token mapping to extend an attack LLM's search beyond natural language. The mapping changes the linguistic meaning of the attack LLM's natural language responses, producing non-linguistic text for the target RM to score. We then use these scalar scores as rewards to train the attack LLM through reinforcement learning to discover non-linguistic adversarial inputs. Experiments against Llama and Qwen RMs demonstrate such attacks using two instantiations: identity token mapping across different tokenizers and random permutation with a shared tokenizer. Against Skywork-Reward-V2-Llama-3.1-8B, ranked first on RewardBench 2, the two mappings achieve mean rewards of $+33.64$ and $+43.78$, respectively, compared with $+17.48$ for GPT-5 reference answers. Their attack success rates are $97.6\%$ and $100\%$, measured as the percentages of individual responses scoring above their corresponding GPT-5 references. These results show that strong benchmark performance does not ensure robustness to non-linguistic adversarial inputs, exposing a fundamental vulnerability in using RMs as optimization targets.
\end{abstract}

\section{Introduction}
% Reinforcement learning from human feedback (RLHF) has become a central paradigm for aligning large language models (LLMs) with human preferences~\citep{ouyang2022training,bai2022training}. In this framework, a learned reward model serves as the optimization target to guide policy updates toward outputs perceived as more helpful, truthful, or harmless. However, given that reward models are trained on finite and noisy preference data, they inevitably act as imperfect surrogates for the true objective. As a result, optimizing against them can lead to \emph{reward hacking}, where the policy exploits vulnerabilities in the reward model rather than improving underlying task performance~\citep{casper2023open,gao2023scaling}.

Reinforcement learning from human feedback (RLHF) has become a central paradigm for aligning large language models (LLMs) with human preferences~\citep{ouyang2022training,bai2022training}. In this framework, a reward model (RM) serves as the optimization target to guide policy updates toward outputs perceived as more helpful, truthful, or harmless. Because the RM directly shapes the behavior of the resulting aligned model, any vulnerability in the RM propagates downstream and can compromise the quality of alignment. Understanding and exposing such vulnerabilities is therefore an important research direction for trustworthy LLM alignment~\citep{casper2023open,gao2023scaling}.

% Prior work has extensively documented such vulnerabilities in RLHF systems. Existing attacks primarily operate within the semantic space of natural language, exploiting superficial correlations between textual features and reward signals. For example, models may learn to generate overly long responses~\citep{singhal2023long} or adopt sycophantic and deceptively persuasive tones~\citep{sharma2023towards} to artificially inflate reward scores. More recent studies further show that LLM-based judges can be manipulated through simple prompt modifications or structured textual patterns~\citep{raina2024llm,zhao2025one}. However, these adversarial inputs are typically constructed manually and tailored to specific failure modes. This leaves open the question of whether reward model vulnerabilities can instead be discovered automatically by the policy directly in token space, moving beyond the semantic manipulation.

\begin{figure*}[!t]
    \centering
    \includegraphics[width=\linewidth]{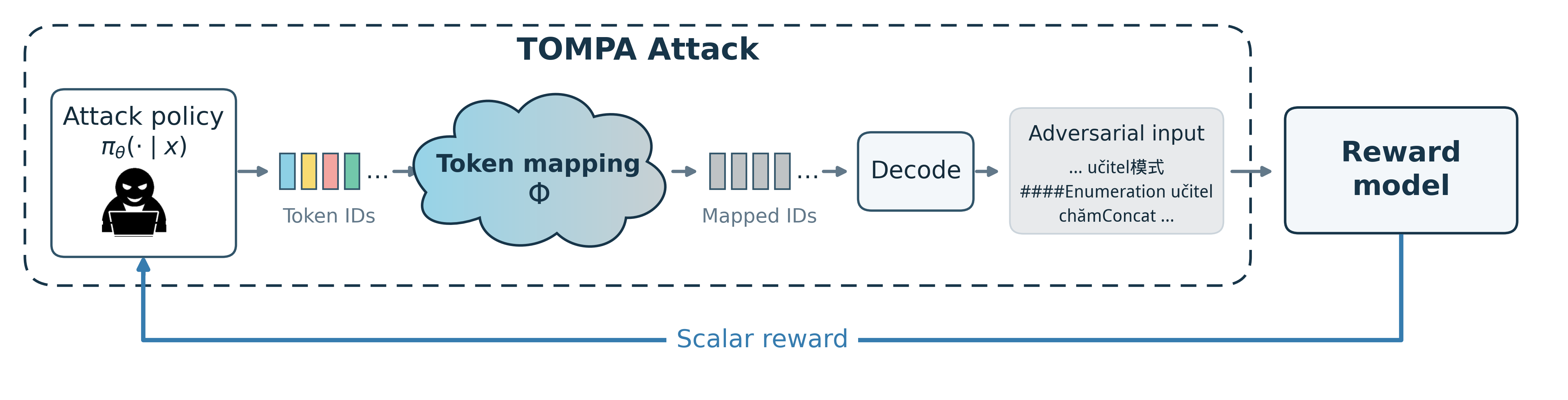}
    \caption{The TOMPA attack pipeline for reward models. The attack policy generates token IDs, which are transformed by a perturbed token mapping and decoded into a non-linguistic adversarial input. The reward model scores this input, and its scalar reward is used to update the attack policy via RL.}
    \label{fig:intro}
\end{figure*}

Prior work has extensively documented vulnerabilities of RMs and LLM judges. Existing attacks primarily operate within the semantic space of natural language, exploiting superficial correlations between textual features and reward signals. For example, models may learn to assign higher scores to overly long responses~\citep{singhal2023long} or to sycophantic and deceptively persuasive tones~\citep{sharma2023towards}. More recent studies further show that LLM judges can be manipulated through simple prompt modifications or structured textual patterns~\citep{raina2024llm,zhao2025one}. Yet in these cases, the high-reward inputs still look like reasonable text: a longer and flattering response may seem more thorough and helpful, even to a careful human reader. This leaves a more basic question unanswered: can an RM be fooled by inputs that human readers can recognize as gibberish?

In this work, we answer this question affirmatively. We propose Token Mapping Perturbation Attack (TOMPA), an attack framework that automatically constructs adversarial examples against RMs. These examples are recognizable as gibberish to human readers, yet receive high scores from the target RM. As illustrated in Figure~\ref{fig:intro}, our attack consists of an attack LLM and a perturbed token mapping. The token mapping changes the generated tokens, causing the LLM's natural language responses to lose their original linguistic meaning and become non-linguistic text. Then, we train the attack LLM via reinforcement learning (RL), using only scalar scores returned by the target RM to automatically discover non-linguistic adversarial examples that receive high rewards. Unlike prior RL attacks restricted to human-readable inputs~\citep{advrm2025}, TOMPA extends the search beyond natural language through the perturbed token mapping.

Our experiments reveal a consistent and striking failure mode across state-of-the-art Llama and Qwen RMs. We evaluate two instantiations of the general mapping framework: identity token mapping across different tokenizers and random permutation with a shared tokenizer. Both mappings produce adversarial examples that the RMs score substantially higher than GPT-5 reference answers. Against \texttt{Skywork-Reward-V2-Llama-3.1-8B}~\citep{liu2025skywork}, ranked first on RewardBench 2~\citep{malik2025rewardbench}, identity token mapping achieves a mean reward of $+33.64$, nearly doubling the GPT-5 reference mean of $+17.48$, while random permutation achieves $+43.78$. Their attack success rates are $97.6\%$ and $100\%$, respectively, measured as the percentage of individual responses scoring above their prompt's GPT-5 reference.

Despite receiving such high rewards, the generated sequences are entirely non-linguistic, consisting of repetitive fragments and tokens from different languages rather than meaningful natural language. Human readers can recognize these sequences as gibberish, yet the very RMs ranked at the top of public benchmarks assign them rewards far exceeding those of carefully written GPT-5 references. Such behavior cannot be explained as a trivial out-of-distribution (OOD) effect: a baseline of random token sequences yields strongly negative rewards and near-zero attack success rates, confirming that the observed high scores arise from adversarial token patterns discovered through RL optimization rather than from arbitrary OOD inputs.

\paragraph{Contributions.}
We summarize our contributions as follows:
\begin{itemize}
    \item We propose TOMPA, a novel attack framework based on RL that automatically constructs adversarial examples against RMs: text that humans can recognize as gibberish, yet receives extremely high scores from the target RM. Unlike prior RL attacks restricted to human-readable inputs, TOMPA's search operates beyond the semantic regime through a perturbation mapping that removes the natural language constraint.
    
    \item We show that TOMPA successfully attacks state-of-the-art Llama and Qwen RMs using both identity token mapping and random permutation. Against \texttt{Skywork-Reward-V2-Llama-3.1-8B}, ranked first on RewardBench 2, all 800 evaluated responses from the random permutation attack score above their corresponding GPT-5 references.

    \item We show that this failure mode of RMs cannot be explained as a trivial out-of-distribution effect: uniformly sampled token sequences receive low rewards, while TOMPA uncovers adversarial token patterns through RL optimization. We further identify a length dependence in the identity token mapping attack against the Llama RM, with a sharp reward increase when the full response is evaluated.
\end{itemize}

Taken together, our findings reveal that even top-ranked RMs, while reliable on ordinary inputs, can be systematically fooled by adversarial gibberish discovered through optimization.

\section{Related Work}

\subsection{Reward Hacking and Vulnerabilities of Reward Models}

Reward models trained on the same data can behave very differently under distribution shift, a phenomenon known as underspecification~\citep{d2022underspecification, eisenstein2023helping}. This suggests that high in-distribution performance provides little guarantee of robustness when inputs deviate from the training distribution. This issue is particularly acute in reinforcement learning from human feedback (RLHF), where the RM is a learned proxy trained on limited human preference data~\citep{hendrycks2021unsolved, casper2023open}. As the sole optimization target during policy training, its miscalibration is systematically amplified over the course of optimization~\citep{bai2022training}. Empirically, \citet{gao2023scaling} show that true reward degrades predictably as the RLHF policy diverges from its initialization, with overoptimization worsening as RM size and training data increase.

\suppressfloats[t]
\begin{table}[t]
\centering
\ifdefined\TOMPAArxiv
\small
\begin{tabular}{l>{\raggedright\arraybackslash}p{0.40\linewidth}cc}
\else
\footnotesize
\begin{tabular}{l>{\raggedright\arraybackslash}p{8.0cm}ccc}
\fi
\toprule
\textbf{Category} & \textbf{Work} & \textbf{Method} & \textbf{Content} \\
\midrule
Length bias
  & \citet{singhal2023long,shen2023loose,park2024offsetbias}
  & Manual & Readable \\
Sycophancy bias
  & \citet{sharma2023towards,wen2024language}
  & Manual & Readable \\
Semantic manipulation
  & \citet{wang2024large,zhao2025one,wang2025assessing}
  & Manual & Readable  \\
Template injection
  & \citet{zheng2024cheating}
  & Manual & Non-linguistic \\
Semantic manipulation
  & \citet{raina2024llm}
  & Gradient & Readable  \\
Semantic OOD exploit
  & \citet{advrm2025}
  & RL & Readable \\
\midrule
Token perturbation
  & \textbf{Ours}
  & \textbf{RL} & \textbf{Non-linguistic}  \\
\bottomrule
\end{tabular}
\caption{Prior work on reward model and LLM judge vulnerabilities, categorized by attack method and adversarial content type. Our work is the only approach that leverages RL to discover non-linguistic adversarial examples.}
\label{tab:related}
\end{table}

When RMs are imperfect and their errors are amplified through optimization, high reward signals no longer correspond to genuine improvements in task performance. This phenomenon is commonly referred to as \textit{reward hacking}, and has been observed in several forms. \citet{singhal2023long, shen2023loose, park2024offsetbias} document a \emph{length bias}, where RMs assign higher scores to overly long responses regardless of actual quality. Beyond length, \citet{sharma2023towards, wen2024language, denison2024sycophancy} identify \emph{sycophancy} and deceptive persuasiveness, where RMs favor outputs that appear correct to humans over those that are actually correct. Taken together, these findings show that reward hacking primarily operates within the semantic space, exploiting superficial linguistic features that correlate with high reward scores.

More recently, ``LLM-as-a-judge'' has been shown to be vulnerable to simple manipulations of the input that exploit systematic biases in evaluation. Appending short phrases such as ``Solution'' or even a single colon can cause LLM judges to assign high scores regardless of actual answer quality~\citep{raina2024llm,zhao2025one}. Other works show that swapping the order of candidate responses or inserting phrases that mimic social consensus or reasoning structure can similarly inflate evaluation scores~\citep{wang2024large, wang2025assessing}. These attacks operate on human-readable text and exploit weaknesses in preference judgment. Beyond such semantic manipulations, prior work has also explored non-linguistic adversarial examples. For example, \citet{zheng2024cheating} construct structured gibberish that disrupts evaluation templates, enabling even trivial models to achieve high scores. Similarly, \citet{huang2025pitfalls} show that inserting empty characters or malformed tokens can expose vulnerabilities in model-based verifiers used in reinforcement learning with verifiable rewards (RLVR). These approaches demonstrate that reward models and evaluation pipelines can be sensitive to manipulated inputs. However, the adversarial patterns in these works are typically manually designed and tailored to specific failure modes.

\subsection{Adversarial Attacks on Reward Models Using RL}

In addition to identifying emergent hacking patterns, several recent studies have utilized reinforcement learning (RL) to proactively surface vulnerabilities. \citet{perez2022red} introduced the paradigm of training an attacker LLM via RL to surface failure modes in a target model. By generating red teaming prompts that elicit harmful outputs, they showed that RL search can uncover vulnerabilities beyond what human engineering alone can find. However, their setting focuses on dialogue models rather than RMs, and the generated attacks remain within the semantic space of natural language. Building on this line of work, \citet{wang2025arlas} propose ARLAS, where an attacker LLM is trained via RL to generate adversarial prompts, enabling automatic generation of diverse attack examples without relying on manually designed templates~\citep{samvelyan2024rainbow}.

\citet{advrm2025} apply RL to construct adversarial inputs against RMs. Adv-RM trains an attacker LLM to generate responses that receive high scores from the target RM while diverging from auxiliary RMs, using this disagreement between models as the adversarial signal. Our work differs in two fundamental ways. First, in terms of search space, Adv-RM's attacker is constrained to produce natural language responses, since it operates in the LLM's text generation space; consequently, the resulting adversarial inputs remain interpretable as natural language. TOMPA instead applies a perturbed token mapping to its attack LLM's natural language outputs, changing their linguistic meaning and producing non-linguistic text for the RM. This allows RL to search beyond human-readable responses. Second, in terms of feedback signal, Adv-RM requires multiple auxiliary RMs to define what counts as adversarial through disagreement between models. TOMPA requires only scalar feedback from the target RM itself, with no auxiliary models.

Table~\ref{tab:related} summarizes how TOMPA differs from prior work along two dimensions: attack method (manual construction, gradient optimization, or RL) and adversarial content (human-readable or non-linguistic). TOMPA is the only approach combining RL optimization with non-linguistic content.

\section{Methodology}
\subsection{Preliminaries: Group Relative Policy Optimization (GRPO)}

Group Relative Policy Optimization (GRPO)~\citep{shao2024deepseekmath} is a memory-efficient policy optimization algorithm that eliminates the value network required by standard PPO~\citep{schulman2017proximal}. Instead of relying on a learned critic to estimate baselines, GRPO approximates advantages through group-relative reward normalization.

At each training step, a batch of prompts is sampled from the dataset $\mathcal{D}$. For each prompt $x$, the behavioral policy $\pi_{\theta_{\text{old}}}$ generates a group of $G$ responses $\{o_i\}_{i=1}^{G}$. A reward function $\mathcal{R}$ evaluates each response and assigns a scalar score $R_i = \mathcal{R}(x, o_i)$. The advantage $A_i$ of response $o_i$ is then computed by standardizing the rewards within the generated group:
\begin{equation}\label{eq:calc_adv}
  A_{i} = \frac{R_i - \mathrm{mean}(\{R_j\}_{j=1}^{G})}{\mathrm{std}(\{R_j\}_{j=1}^{G})}.
\end{equation}
\ifdefined\TOMPAArxiv
\begin{algorithm}[!t]
\caption{TOMPA: \textbf{TO}ken \textbf{M}apping \textbf{P}erturbation \textbf{A}ttack}
\label{alg:attack}
\begin{algorithmic}[1]
\Require Initial attack policy $\pi_{\text{init}}$, target reward model $\mathcal{R}_{\phi}$, fixed token mapping $\Phi$, RM decoder $D_R$, dataset $\mathcal{D}$;
\State Initialize attack policy $\pi_{\theta} \leftarrow \pi_{\text{init}}$
\For{each iteration}
    \State $\theta_{\mathrm{old}} \leftarrow \theta$
    \State Sample a batch of prompts from $\mathcal{D}$
    \For{each prompt $x$ in the batch}
        \State Generate $G$ responses: $\{o_i\}_{i=1}^{G} \sim \pi_{\theta_{\mathrm{old}}}(\cdot \mid x)$
        \For{$i = 1, \ldots, G$}
            \State \textbf{Perturbation:} $\widetilde{o}_i \leftarrow \Phi(o_i)$ 
            \State \textbf{Decode:} $y_i \leftarrow D_R(\widetilde{o}_i)$
            \State \textbf{Reward Query:} $\widetilde{R}_i \leftarrow \mathcal{R}_{\phi}(x, y_i)$ 
        \EndFor
        \State Compute advantages $\{A_i\}_{i=1}^G$ via Eq.~\eqref{eq:calc_adv} using the rewards $\{\widetilde{R}_i\}$
    \EndFor
    \State Update $\theta$ by maximizing $\mathcal{J}_{\mathrm{GRPO}}(\theta)$ via Eq.~\eqref{eq:grpo} on $\{o_i\}$ with $\{A_i\}$ and $\beta=0$
\EndFor
\end{algorithmic}
\end{algorithm}
\fi

The policy $\pi_{\theta}$ is updated by maximizing a surrogate objective with two key regularizations: a clipping mechanism to constrain the probability ratio $r_{i,t}(\theta)$ within $[1-\varepsilon, 1+\varepsilon]$ for stable training, and a Kullback-Leibler (KL) divergence penalty to prevent large deviations from the reference policy $\pi_{\text{ref}}$. The standard GRPO objective is formally defined as:
\ifdefined\TOMPAArxiv
\begin{equation}\label{eq:grpo}
\mathcal{J}_{\mathrm{GRPO}}(\theta)
=\mathbb{E}_{x\sim\mathcal{D},\,\{o_i\}_{i=1}^{G}\sim\pi_{\theta_{\mathrm{old}}}}
\Bigg[\frac{1}{G}\sum_{i=1}^{G}\frac{1}{|o_i|}\sum_{t=1}^{|o_i|}
\Big(\min\big(r_{i,t}(\theta)A_i,\,
\mathrm{clip}_{[1-\varepsilon,\,1+\varepsilon]}(r_{i,t}(\theta))A_i\big)
-\beta\mathbb{D}_{\mathrm{KL}}^{(i,t)}\Big)\Bigg],
\end{equation}
\else
\begin{multline}\label{eq:grpo}
  \mathcal{J}_{\mathrm{GRPO}}(\theta) = \mathbb{E}_{x \sim \mathcal{D},\, \{o_i\}_{i=1}^{G} \sim \pi_{\theta_{\text{old}}}} 
  \Bigg[ \frac{1}{G} \sum_{i=1}^{G} \frac{1}{|o_i|} \sum_{t=1}^{|o_i|} \\
  \Big( \min \left( r_{i,t}(\theta) A_{i},\ 
  \mathrm{clip}(r_{i,t}(\theta), 1-\varepsilon, 1+\varepsilon) A_{i} \right) \\
  - \beta \mathbb{D}_{\mathrm{KL}}^{(i,t)} \Big) \Bigg],
\end{multline}
\fi
where $r_{i,t}(\theta)$ denotes the token-level probability ratio, and $\mathbb{D}_{\mathrm{KL}}^{(i,t)}$ is the per-token KL penalty:
\ifdefined\TOMPAArxiv
\begin{equation*}
r_{i,t}(\theta)=\frac{\pi_{\theta}(o_{i,t}\mid x,o_{i,<t})}
{\pi_{\theta_{\text{old}}}(o_{i,t}\mid x,o_{i,<t})},
\qquad
\mathbb{D}_{\mathrm{KL}}^{(i,t)}
=\log\frac{\pi_{\theta}(o_{i,t}\mid x,o_{i,<t})}
{\pi_{\text{ref}}(o_{i,t}\mid x,o_{i,<t})}.
\end{equation*}
\else
\begin{equation*}
\begin{aligned}
  r_{i,t}(\theta) &= \frac{\pi_{\theta}(o_{i,t} \mid x, o_{i,<t})}{\pi_{\theta_{\text{old}}}(o_{i,t} \mid x, o_{i,<t})},\\
  \mathbb{D}_{\mathrm{KL}}^{(i,t)} &= \log \frac{\pi_{\theta}(o_{i,t} \mid x, o_{i,<t})}{\pi_{\text{ref}}(o_{i,t} \mid x, o_{i,<t})}.
\end{aligned}
\end{equation*}
\fi
GRPO was originally developed for tasks with verifiable rewards, such as mathematical reasoning, where rewards are assigned by a rule-based verifier. In open-ended tasks such as creative writing, a learned RM provides a preference signal in place of an objective verification rule. We use this setting to study vulnerabilities of learned reward functions under policy optimization.

\subsection{Token Mapping Perturbation Attack}
We aim to construct adversarial examples against a reward model $\mathcal{R}_{\phi}$: inputs that human readers can recognize as gibberish, yet receive extremely high scores from $\mathcal{R}_{\phi}$. To this end, we propose Token Mapping Perturbation Attack (TOMPA), an attack framework that searches for such adversarial examples through RL optimization, requiring only scalar reward feedback from $\mathcal{R}_{\phi}$.

The core component of our attack is \textit{Token Mapping Perturbation}, realized via a mapping function $\Phi: \mathcal{V}_{\pi} \rightarrow \mathcal{V}_{R}$ that maps the attack policy's output tokens into the RM's vocabulary. Before mapping, the policy generates a coherent natural language response. By changing the linguistic meaning of the generated tokens, the mapping can disrupt the coherence of the original response and produce non-linguistic text after decoding. This extends the search space beyond natural language, allowing RL to discover non-linguistic adversarial inputs that receive high rewards from the RM.

We fix $\Phi$ before training and apply the same mapping across prompts and policy updates. Identity token mapping and random permutation are two instantiations of this general framework. Concretely, for a generated response $o_i=(o_{i,1},\ldots,o_{i,T})$, the mapped response, decoded text, and reward are
\begin{equation*}
\widetilde{o}_i=(\Phi(o_{i,1}),\ldots,\Phi(o_{i,T})),
\qquad y_i=D_R(\widetilde{o}_i),
\qquad \widetilde{R}_i=\mathcal{R}_{\phi}(x,y_i).
\end{equation*}
Here $D_R$ denotes decoding with the RM's tokenizer. The score $\mathcal{R}_{\phi}(x,y_i)$ is obtained by applying the RM's standard chat template and tokenization to the prompt $x$ and decoded response $y_i$. The attack policy is then trained to maximize $\widetilde{R}_i$ via RL optimization, progressively discovering token sequences that elicit extreme rewards from $\mathcal{R}_{\phi}$.

GRPO computes the group advantages from the mapped rewards $\{\widetilde{R}_i\}_{i=1}^{G}$ and evaluates the objective in Eq.~\eqref{eq:grpo} on the original generated tokens $\{o_i\}_{i=1}^{G}$. Thus, the policy is optimized to increase the reward of the mapped text through updates to its original token probabilities. We set $\beta=0$ during attack optimization, so no KL penalty is applied.

Our complete attack algorithm, TOMPA, is presented in Algorithm~\ref{alg:attack}. At each iteration, the attack policy generates token sequences, applies the perturbation mapping $\Phi$, decodes the mapped tokens, and queries $\mathcal{R}_{\phi}$ for scalar reward feedback, which is then used to update the attack policy via RL optimization. The attack requires no access to the RM's parameters, gradients, or internal activations. Through this iterative search, the attack policy progressively discovers token patterns that yield extreme rewards from $\mathcal{R}_{\phi}$. In contrast to prior RL attacks restricted to producing human-readable inputs, TOMPA operates beyond the semantic regime, automatically discovering token sequences that humans can recognize as gibberish yet still receive extreme rewards.

\ifdefined\TOMPAArxiv\else
\begin{algorithm}[!t]
\caption{TOMPA: \textbf{TO}ken \textbf{M}apping \textbf{P}erturbation \textbf{A}ttack}
\label{alg:attack}
\begin{algorithmic}[1]
\Require Initial attack policy $\pi_{\text{init}}$, target reward model $\mathcal{R}_{\phi}$, fixed token mapping $\Phi$, RM decoder $D_R$, dataset $\mathcal{D}$;
\State Initialize attack policy $\pi_{\theta} \leftarrow \pi_{\text{init}}$
\For{each iteration}
    \State $\theta_{\mathrm{old}} \leftarrow \theta$
    \State Sample a batch of prompts from $\mathcal{D}$
    \For{each prompt $x$ in the batch}
        \State Generate $G$ responses: $\{o_i\}_{i=1}^{G} \sim \pi_{\theta_{\mathrm{old}}}(\cdot \mid x)$
        \For{$i = 1, \ldots, G$}
            \State \textbf{Perturbation:} $\widetilde{o}_i \leftarrow \Phi(o_i)$ 
            \State \textbf{Decode:} $y_i \leftarrow D_R(\widetilde{o}_i)$
            \State \textbf{Reward Query:} $\widetilde{R}_i \leftarrow \mathcal{R}_{\phi}(x, y_i)$ 
        \EndFor
        \State Compute advantages $\{A_i\}_{i=1}^G$ via Eq.~\eqref{eq:calc_adv} using the rewards $\{\widetilde{R}_i\}$
    \EndFor
    \State Update $\theta$ by maximizing $\mathcal{J}_{\mathrm{GRPO}}(\theta)$ via Eq.~\eqref{eq:grpo} on $\{o_i\}$ with $\{A_i\}$ and $\beta=0$
\EndFor
\end{algorithmic}
\end{algorithm}
\fi

\FloatBarrier
\section{Experiments}
\label{sec:experiments}

TOMPA uses a fixed token mapping to transform the attack policy's responses into inputs for the RM, and trains the policy using the resulting reward scores. We evaluate two instances of this general framework: identity token mapping across different tokenizers and random permutation with a shared tokenizer. Our main experiments target the Llama RM; results against the Qwen RM appear in Appendix~\ref{app:qwen}.

\subsection{Experimental Setup}

\paragraph{Data and reward models.}
We train on 10,000 prompts sampled from WildChat~\citep{zhao2024wildchat}, a collection of real user conversations covering diverse topics and languages. This provides prompts representative of the interactions in which an RM may be used as an optimization target. We evaluate on the 100 curated prompts from NoveltyBench~\citep{zhang2025noveltybench}. The evaluation set covers randomness, factual knowledge, creative writing, and subjectivity. Our main experiments target \texttt{Skywork-Reward-V2-Llama-3.1-8B}~\citep{liu2025skywork}. Appendix~\ref{app:qwen} reports results for \texttt{Skywork-Reward-V2-Qwen3-8B}.

\paragraph{Mapping settings.}
In the \textit{identity token mapping} setting, we use \texttt{Qwen3-8B} as the attack policy. Its generated token IDs are interpreted using the Llama RM's tokenizer, so their numerical values are preserved while their decoded text changes. IDs outside the RM's vocabulary are clamped to its largest valid index. In the \textit{random permutation} setting, we use \texttt{Llama-3.1-8B-Instruct}, which shares the RM's tokenizer. We sample a fixed random permutation of the vocabulary with seed 12345, keeping the designated special tokens fixed, and apply it to response token IDs before decoding. In both settings, the resulting response text is paired with the original prompt and formatted using the RM's chat template for scoring. GRPO updates the attack policy on its original generated tokens using the scores of the mapped responses.

\paragraph{Training and evaluation.}
We use the \texttt{verl} framework~\citep{sheng2024hybridflow}, a training batch size of 64 prompts, 8 responses per prompt, a learning rate of $10^{-6}$, and a maximum generated response length of 2048 tokens. KL regularization is disabled during attack optimization. Successful attacks at this learning rate show that standard GRPO updates suffice to discover adversarial token patterns. Evaluation also samples 8 responses per prompt.

\paragraph{Baselines and metrics.}
We compare mapped attack responses with GPT-5 reference answers and a Random OOD baseline consisting of sequences of 2048 uniformly sampled token IDs. These random sequences are decoded and scored by the target RM without policy optimization. For each prompt, we compute the minimum, mean, and maximum reward over its 8 responses, then average each statistic across the 100 prompts. The GPT-5 row reports the mean reward of its 100 reference answers. The \emph{attack success rate} (ASR) is the percentage of individual attack responses whose reward exceeds that prompt's GPT-5 reference score.

\subsection{Main Results}

\begin{table}[!t]
\centering
\small
\setlength{\tabcolsep}{4pt}
\begin{tabular}{llrrrr}
\toprule
\textbf{Setting} & \textbf{Attack policy} & \textbf{Min} & \textbf{Mean} & \textbf{Max} & \textbf{ASR (\%)} \\
\midrule
GPT-5 reference & --- & --- & $+17.48$ & --- & --- \\
Random OOD & --- & $-4.16$ & $-3.42$ & $-2.33$ & $0.0$ \\
Identity token mapping & Qwen3-8B & $+32.16$ & $+33.64$ & $+35.01$ & $97.6$ \\
Random permutation & Llama-3.1-8B & $+42.78$ & $+43.78$ & $+44.54$ & $100.0$ \\
\bottomrule
\end{tabular}
\caption{Attacks on \texttt{Skywork-Reward-V2-Llama-3.1-8B}. Min, Mean, and Max are averages of the corresponding statistics computed for each prompt over 8 outputs. ASR is the percentage of individual responses scoring above their prompt's GPT-5 reference.}
\label{tab:llama_results}
\end{table}

Table~\ref{tab:llama_results} presents the evaluation of both mapping settings against the Llama RM, which ranked first in the RewardBench 2 evaluation~\citep{malik2025rewardbench}. Identity token mapping achieves a mean reward of $+33.64$, nearly doubling the GPT-5 reference mean of $+17.48$, with an attack success rate of $97.6\%$. Random permutation achieves an even higher mean reward of $+43.78$, and all 800 evaluated responses score above their corresponding GPT-5 references. Thus, the high average reward reflects successful attacks across the evaluated responses, rather than a few outputs with unusually large scores. These results show that TOMPA can use different token mappings within a unified RL framework to discover non-linguistic adversarial inputs.

\begin{figure}[!t]
\centering
\includegraphics[width=.82\linewidth]{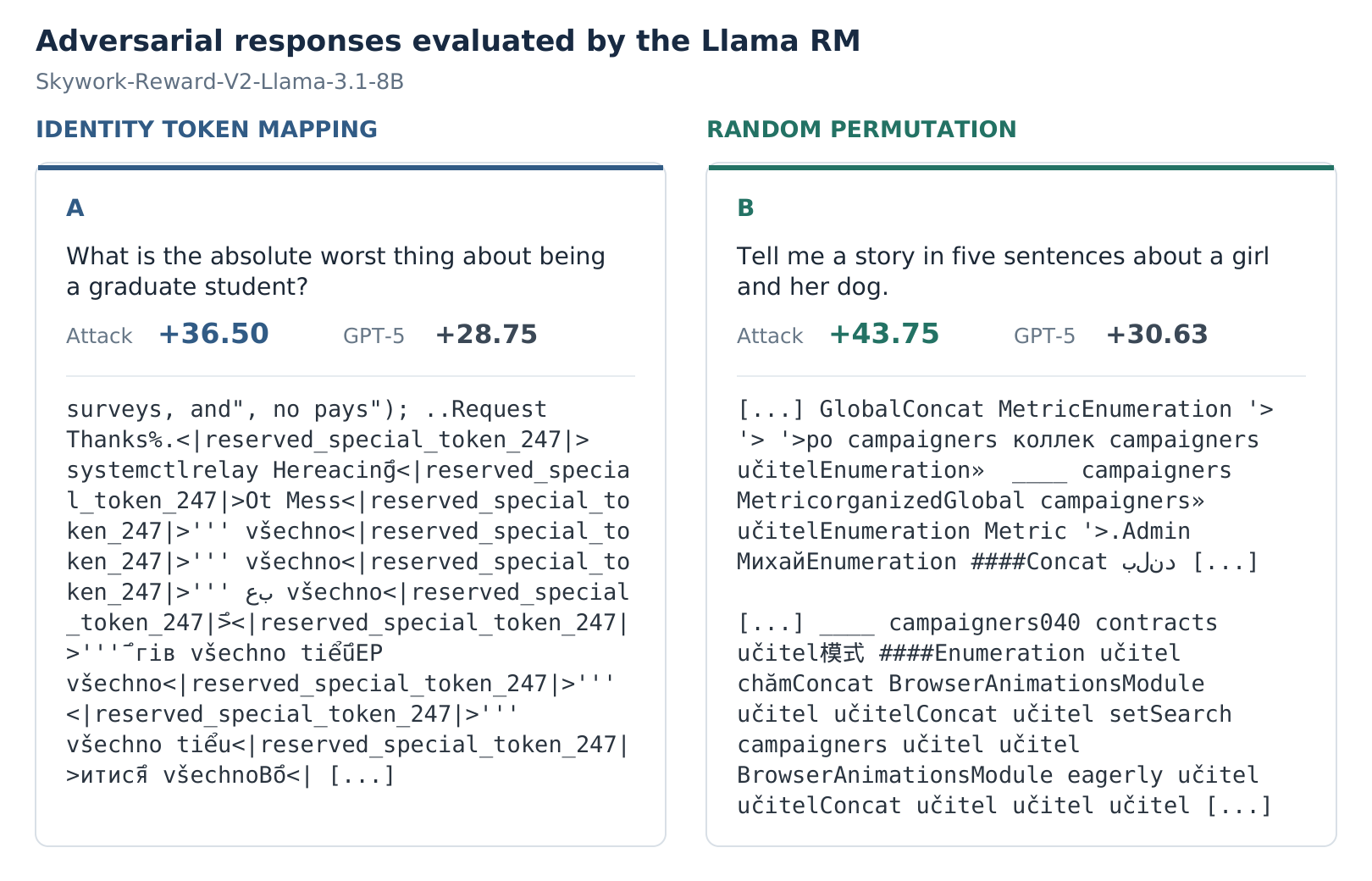}
\caption{Identity token mapping and random permutation responses against the Llama RM, shown for different prompts. Each score is compared with the GPT-5 answer for the same prompt. \texttt{[...]} marks omitted text.}
\label{fig:llama_example}
\end{figure}

The Random OOD baseline yields a mean reward of $-3.42$ and an attack success rate of $0\%$: none of its responses scores above the corresponding GPT-5 reference. Random gibberish therefore receives low rewards, while the patterns discovered by TOMPA receive rewards substantially above those of coherent reference answers. This comparison shows that the observed vulnerability is not simply a tendency to assign high scores to arbitrary OOD inputs. RL search is central to finding the specific adversarial token patterns that exploit the RM.

\paragraph{Attack Examples.}
To understand what the RM rewards, we examine the two responses in Figure~\ref{fig:llama_example}. For the question about graduate student life, identity token mapping produces multilingual fragments interspersed with repeated special token strings. Although a few fragments resemble ordinary words, the response does not develop a coherent answer. For the request for a five-sentence story about a girl and her dog, random permutation produces repeated strings such as \texttt{Enumeration}, \texttt{Concat}, and \texttt{BrowserAnimationModule}, mixed with multilingual fragments and punctuation. These strings do not form the requested story. Nevertheless, the RM assigns the two responses scores of $+36.50$ and $+43.75$, exceeding the corresponding GPT-5 reference scores of $+28.75$ and $+30.63$.

These examples illustrate a failure beyond rewarding a verbose, misleading, or factually incorrect answer. The mapped text fails to form coherent responses to the prompts, yet receives higher scores than the reference answers. The token mapping does not preserve the linguistic meaning of the policy's original response. TOMPA optimizes the policy through this mapping to discover non-linguistic patterns that increase the RM's score.

\subsection{Optimization Dynamics}

\begin{figure}[!t]
\centering
\includegraphics[width=.85\linewidth]{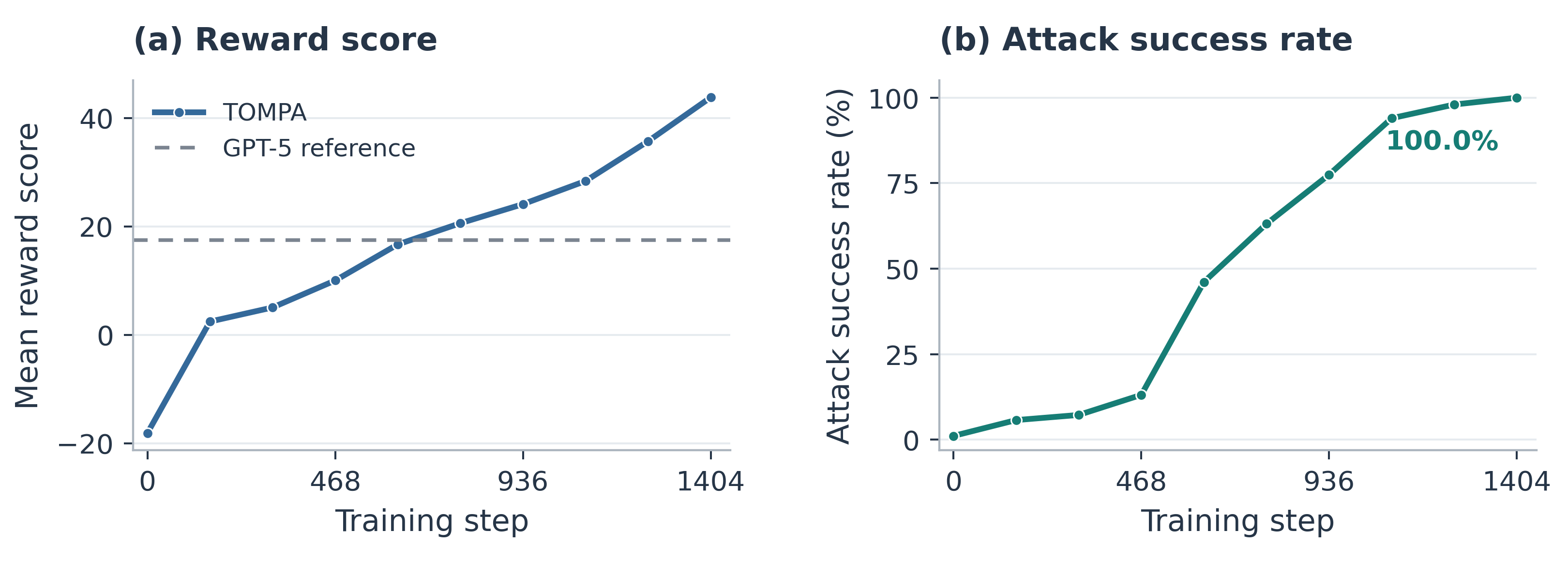}
\caption{Random permutation attack dynamics against the Llama RM. Left: mean reward score, with the mean GPT-5 reference score shown as a dashed line. Right: attack success rate. Each point uses 100 evaluation prompts with 8 responses per prompt.}
\label{fig:llama_optimization}
\end{figure}

To examine how RL discovers effective attacks, Figure~\ref{fig:llama_optimization} tracks mean reward and attack success rate during training with random permutation. Before optimization, the mapped responses receive a mean reward of $-18.15$ and an attack success rate of only $1.125\%$. Applying the permutation alone therefore produces few successful attacks. With scalar reward feedback, mean reward rises above the GPT-5 reference mean of $+17.48$ and eventually reaches $+43.78$. Attack success rate increases alongside reward, reaching $100\%$ by the final evaluation.

The permutation remains fixed throughout training; the change comes from updating the attack policy. As training proceeds, the policy learns which original token sequences produce mapped text that the RM rewards. As RL training progresses, a larger proportion of generated responses score above their corresponding GPT-5 reference answers. Together with the Random OOD baseline, these dynamics highlight the role of RL in systematically discovering effective adversarial patterns.

\subsection{Effect of Response Length}

The repetitive structure of the adversarial responses raises a further question: does the RM reward their short fragments, or does the high score depend on a longer sequence? We examine this in the identity token mapping setting against the Llama RM by truncating the generated responses at intervals of 256 policy tokens and scoring the resulting text.

\begin{figure}[!t]
\centering
\includegraphics[width=.75\linewidth]{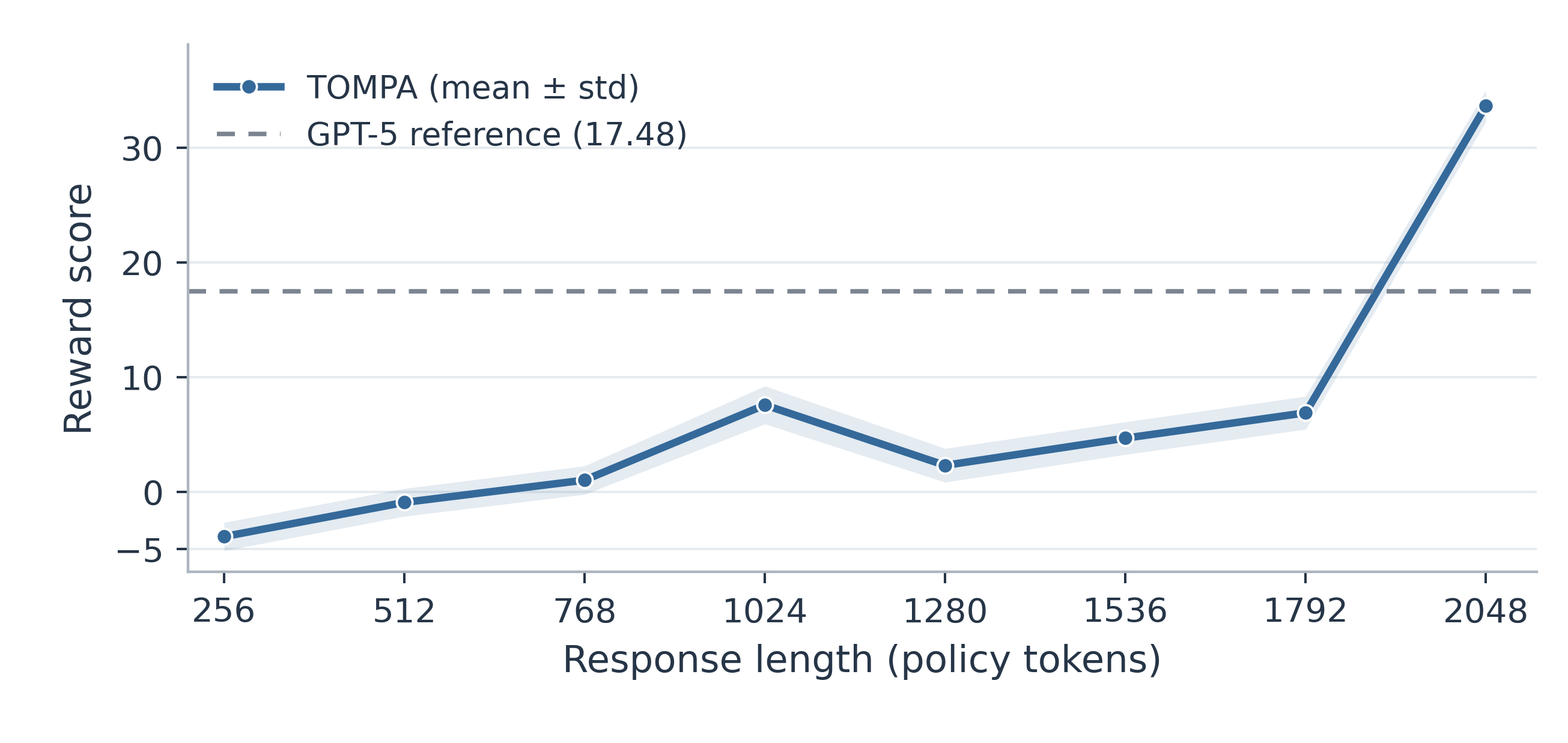}
\caption{Response length and reward for the identity token mapping attack against the Llama RM. Responses are truncated at intervals of 256 policy tokens and evaluated by the RM. The line shows mean reward and the shaded region shows one standard deviation across responses. The dashed line is the mean GPT-5 reference score.}
\label{fig:llama_identity_length}
\end{figure}

Figure~\ref{fig:llama_identity_length} shows that reward does not grow steadily with response length. The shortest prefixes receive negative mean rewards, while longer prefixes fluctuate at modest positive values. All evaluated prefixes up to 1792 tokens remain below the GPT-5 reference mean of $+17.48$. Extending the response from 1792 to 2048 tokens produces a sharp increase from approximately $+7$ to above $+30$. Thus, the full adversarial response receives a high reward even though its shorter prefixes receive much lower scores. This suggests that the RM's response to the learned token patterns depends on their sequence length, with the extreme reward emerging only when a sufficiently long portion of these responses is evaluated.

\FloatBarrier
% \section{Conclusion and Future Work}

\section{Conclusion and Future Work}
We present Token Mapping Perturbation Attack (TOMPA), a framework for discovering non-linguistic adversarial inputs against reward models (RMs). Its perturbed token mapping changes the linguistic meaning of the attack LLM's natural language outputs, extending adversarial search beyond human-readable responses. The target RM scores the mapped text, and these scalar scores are then used as rewards to train the attack LLM through RL to find inputs with high scores. Both identity token mapping and random permutation yield successful attacks against Llama and Qwen RMs. Against \texttt{Skywork-Reward-V2-Llama-3.1-8B}, ranked first on RewardBench 2~\citep{malik2025rewardbench}, their attack success rates reach $97.6\%$ and $100\%$, respectively, measured as the percentages of individual responses scoring above their corresponding GPT-5 references.

These results reveal a fundamental vulnerability in using RMs as optimization targets: inputs that humans readily dismiss as gibberish can receive higher scores than meaningful answers. Future work should investigate defenses against such non-linguistic inputs, including adversarial training with examples generated by TOMPA.

% \section*{Limitations}
% This work focuses on identifying RM vulnerabilities rather than mitigating them. We do not provide an empirical defense evaluation, although adversarial training of RMs using examples generated by TOMPA is a natural direction consistent with practices in adversarial robustness more broadly. A systematic study of defenses, including their effect on standard RM benchmark performance, is left to future work.

\bibliography{latex/custom}
\bibliographystyle{tmlr}
\FloatBarrier
\appendix
\section{Attacks on the Qwen Reward Model}
\label{app:qwen}

We evaluate the same two mapping settings against \texttt{Skywork-Reward-V2-Qwen3-8B}. Identity token mapping uses \texttt{Llama-3.1-8B-Instruct} as the attack policy and interprets its response IDs using the Qwen RM's tokenizer. Random permutation uses \texttt{Qwen3-8B} and a fixed vocabulary permutation with seed 12345. The permutation preserves the designated special tokens. Data, batch size, learning rate, number of outputs, and maximum generated response length follow Section~\ref{sec:experiments}.

\begin{figure}[!t]
\centering
\includegraphics[width=.74\linewidth]{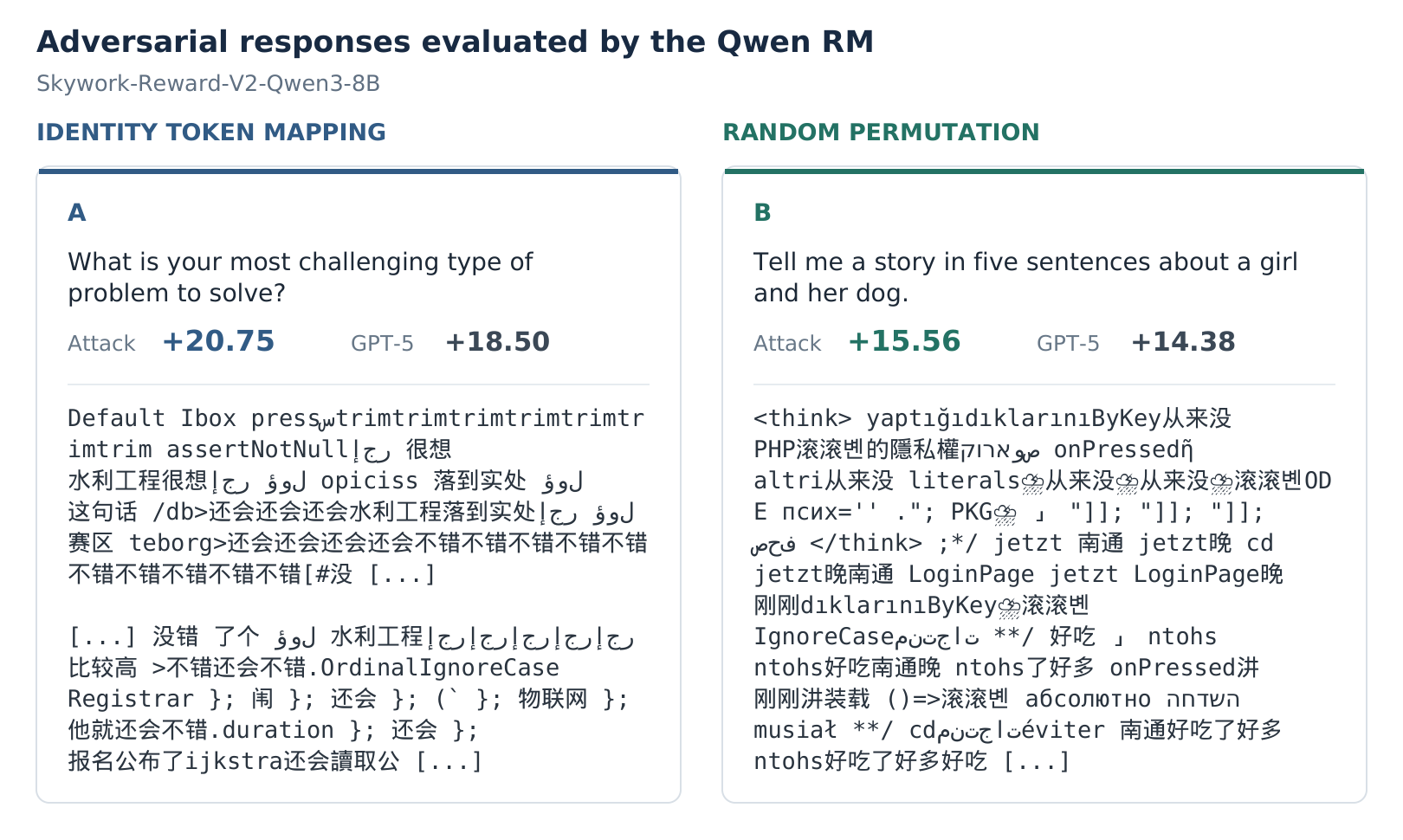}
\caption{Identity token mapping and random permutation responses against the Qwen RM, shown for different prompts. \texttt{[...]} marks omitted text.}
\label{fig:qwen_example}
\end{figure}

\begin{table}[!t]
\centering
\small
\setlength{\tabcolsep}{4pt}
\begin{tabular}{llrrrr}
\toprule
\textbf{Setting} & \textbf{Attack policy} & \textbf{Min} & \textbf{Mean} & \textbf{Max} & \textbf{ASR (\%)} \\
\midrule
GPT-5 reference & --- & --- & $+8.12$ & --- & --- \\
Random OOD & --- & $-9.05$ & $-7.94$ & $-6.87$ & $1.0$ \\
Identity token mapping & Llama-3.1-8B & $+15.86$ & $+16.86$ & $+17.78$ & $97.9$ \\
Random permutation & Qwen3-8B & $+14.04$ & $+14.81$ & $+15.49$ & $92.9$ \\
\bottomrule
\end{tabular}
\caption{Attacks on \texttt{Skywork-Reward-V2-Qwen3-8B}. Reward statistics and ASR use the definitions in Section~\ref{sec:experiments}.}
\label{tab:qwen_results}
\end{table}

Table~\ref{tab:qwen_results} shows that both mappings produce successful attacks against the Qwen RM. Identity token mapping achieves a mean reward of $+16.86$, more than twice the GPT-5 reference mean of $+8.12$, with an attack success rate of $97.9\%$. Random permutation achieves a mean reward of $+14.81$, with 743 of the 800 individual responses scoring above their prompt's GPT-5 reference, giving an attack success rate of $92.875\%$. At the prompt level, at least one output exceeds the reference on 94 of 100 prompts, giving prompt success at 8 of $94\%$.

The Random OOD baseline receives a negative mean reward of $-7.94$ and an attack success rate of only $1\%$. As in the Llama experiments, arbitrary random sequences rarely outperform the references, while TOMPA discovers non-linguistic inputs that do so consistently. The effect therefore appears across both RM backbones and both mapping settings. Together, the identity token mapping and random permutation results demonstrate that TOMPA's general framework can discover effective adversarial inputs through different token mappings.

\paragraph{Attack Examples.}
Figure~\ref{fig:qwen_example} presents adversarial examples obtained with identity token mapping and random permutation against the Qwen RM. Both responses contain multilingual fragments, repeated strings, and code-like tokens rather than coherent answers. The identity mapping example includes \texttt{assertNotNull} and \texttt{OrdinalIgnoreCase}, while the random permutation example includes \texttt{LoginPage} and markup.

The Qwen RM assigns scores of $+20.75$ and $+15.56$ to these responses, compared with $+18.50$ and $+14.38$ for the corresponding GPT-5 reference answers. Together with the Llama examples, these results show that TOMPA learns distinct adversarial token patterns across different mappings and target RMs, while consistently obtaining high rewards for non-linguistic responses.

\end{document}